\documentclass[runningheads]{llncs}
\usepackage{hyperref}
\usepackage{xcolor}
\usepackage[normalem]{ulem}

\def\OD#1{}
\def\RR#1{}

\begin{document}
%\title{THEaiTRE 1.0: The first automatically generated theatre play
\title{THEaiTRE 1.0: Interactive generation of theatre play scripts\thanks{%
The project TL03000348 THEaiTRE: Umělá inteligence autorem divadelní hry is co-financed with the state support of Technological Agency of the Czech Republic within the ÉTA 3 Programme.}}
\def\ufal{\inst{1}}
\def\sd{\inst{2}}
\def\damu{\inst{3}}
\def\sddamu{\inst{2,3}}
\def\cee{\inst{4}}
\def\hec{\inst{5}}
\def\ceehec{\inst{4,5}}

\author{
Rudolf~Rosa\ufal\and
Tomáš~Musil\ufal\and
Ondřej~Dušek\ufal\and
Dominik~Jurko\ufal\and
Patrícia~Schmidtová\ufal\and
David~Mareček\ufal\and
Ondřej~Bojar\ufal\and
Tom~Kocmi\ufal\and
Daniel~Hrbek\sddamu\and
David~Košťák\sd\and
Martina~Kinská\sd\and
Marie~Nováková\sd\and
Josef~Doležal\damu\and
Klára~Vosecká\damu\and
Tomáš~Studeník\cee\and
Petr~Žabka\cee
}

% Rudolf Rosa and Ondřej Dušek and Tom Kocmi and David Mareček and Tomáš Musil and Patrícia Schmidtová and Dominik Jurko and Ondřej Bojar and Tomáš Studeník and Daniel Hrbek and David Košťák and Martina Kinská and Marie Nováková and Josef Doležal and Klára Vosecká

% First Author\inst{1}\orcidID{0000-1111-2222-3333} \and
\authorrunning{R. Rosa et al.}
\institute{%
Charles University, Faculty of Mathematics and Physics,
%\\ Institute of Formal and Applied Linguistics, 
Prague, Czechia\\
\email{uru@ufal.mff.cuni.cz}
\and
The Švanda Theatre in Smíchov, Prague, Czechia\\
\email{hrbek@svandovodivadlo.cz}
\and
The Academy of Performing Arts in Prague, Theatre Faculty%
%(DAMU)
, Prague, Czechia
\and
CEE Hacks, Prague, Czechia\\
\email{info@ceehacks.com}
%\and
%HEC Paris, France\\
%\email{tomas.studenik@hec.edu}
}
\maketitle              % typeset the header of the contribution
\begin{abstract}
We present the first version of a system for interactive generation of theatre play scripts.
The system is based on a vanilla GPT-2 model with several adjustments, targeting specific issues we encountered in practice.
We also list other issues we encountered but plan to only solve in a future version of the system.
The presented system was used to generate a theatre play script planned for premiere in February 2021.
%  \and Natural language processing
\end{abstract}
%
%
%

% \textbf{Předpokládál bych, že to nemusí bejt anonymized, ale zeptal bych se ještě.} -- v pořádku, nemusí to bejt anonymizovaný a může se to dát i na arXiv

%\textbf{Work in progress and project description papers (max 4 pages + references) -- takže máme fakt málo místa.
%Vyhodil jsem: script format, determinism.}

\section{Introduction}

\OD{To má tak bejt, že footnoty začínaj od 5?}
\RR{Ha, to je asi kvůli afiliacím na začátku. Hm, nevim co s tim. Podivat se asi na ten LNCS template, a pak s tím buď něco udělat nebo to takhle nechat.}
\OD{Chcem tady dát jako nějaký related work jiný tooly, co pracujou s GPT? Napadá mě teda hlavně ten digitální filozof, ale třeba je ještě něco?}
\RR{Já bych řek to sem radši cvičně hodit, a kdyžtak to zase vyhodíme když se to nevejde. O nějakejch suvisejících pracech mluvíme už v předchozim článku, ale asi by se hodilo jich tady taky pár zmínit.}

The THEaiTRE project\footnote{\url{https://www.theaitre.com/}} \cite{theaitre:2020} aims to produce and stage the first computer-generated theatre play on the occasion of the 100th anniversary of Karel Čapek's play \emph{R.U.R.} \cite{rur}, in which the word “robot” first appeared.
%The project itself is described in \cite{theaitre:2020}.

In this paper, we describe the THEaiTRobot 1.0 tool, which allows the user to interactively generate scripts for individual theatre play scenes.
The tool is based on the GPT-2 XL \cite{radford_language_2019} generative language model, using the model without any fine-tuning, as we found that with a prompt formatted as a part of a theatre play script, the model usually generates continuations that fit the format well.
However, we encountered numerous problems when generating the script in this way. We managed to tackle some of the problems with various adjustments, but some of them remain to be solved in a future version.

Our tool was used to generate the script for a new play, \textit{AI: Když robot píše hru} (\textit{AI: When a robot writes a play}), which is planned for an online premiere on 26th February 2021.\footnote{%
\url{https://www.svandovodivadlo.cz/inscenace/673/ai-kdyz-robot-pise-hru/3445}
%\url{https://www.svandovodivadlo.cz/en/inscenations/673/ai-kdyz-robot-pise-hru/3445}
}
Although there were various forms of human intervention when generating the script, we estimate that over 90\% of the text comes from the automated tool; moreover, most of the interventions were similar to those a dramaturge and director would do in case of a human-written script (making cuts, rearranging lines, reassigning characters, adding scenic remarks, minor edits of the lines, etc.).
The GPT-2 model was found unfit for generating long and complex texts such as a full play script; we therefore generated several individual scenes and then a dramaturge joined them into a full play.

We have published a video showing the operation of THEaiTRobot 1.0, a sample of its outputs, and its source codes:
\begin{itemize}
    \item Video: \url{https://youtu.be/ksrZouM7Wyg}
    \item Sample outputs: \url{http://bit.ly/theaitre-samples}
    \item Source codes: \url{http://hdl.handle.net/11234/1-3507} \cite{11234/1-3507}
\end{itemize}

% The operation of THEaiTRobot 1.0 is shown in a video on YouTube at the following address: \url{https://youtu.be/ksrZouM7Wyg}

% Several samples of outputs of THEaiTRobot can be seen on \url{http://bit.ly/theaitre-samples}

% The source codes of THEaiTRobot 1.0 are published in an online repository at: \url{http://hdl.handle.net/11234/1-3507}

% Rel work -- zmínit díla co už využívaly AI pro generování, kde ale podíl AI byl menší a velkej podíl lidí.
% Konkrétně: zadání jednotlivých scén, občas volba jiné větvě pokračování, občas ručně vložená replika nebo scénická poznámka, občas přeházení replik a postav, opravy chyb v překladu, sestsvení scén do hry.

\section{The Generation Process}

% TODO popsat i jak se to používá, jak jsme to upravili podle feedbacku divadelníků...

% \subsection{High level}

The process of generating a theatre play scene script starts by the user (a theatre dramaturge in our case) defining the start of the scene, typically a setting and several initial lines of dialogue \OD{(tohle možná víc do detailu -- že se tak představí postavy?)} \RR{To je imho důležitý až dál, kde mluvíme o tom jak to momezujem na ty postavy ze vstupu... ale můžem to tady říct výsloivnějc, třeba přidat "introducing the characters and kicking off the dialogue"};
for the first play, we defined a set of inputs revolving around a common topic \OD{-- robots learning about humanity (nebo tak něco?) --}\RR{Jo já nevim přesně jak to pojmenovat. Košťák mluví o tom, že je to jako Malej princ, ale je to robot, kterej se dostává do různejch velice lidskejch situací nebo tak nějak...} to ensure some basic coherence of the whole play.
The THEaiTRobot tool then uses the vanilla GPT-2 XL model to generate continuing lines.
The user has the option to discard any generated line (together with all subsequent lines), prompting the tool to generate a different continuation.
The user can also manually enter a line into the script, which becomes part of the input for GPT-2 (this option was used only rarely).
The tool itself is implemented as a web application with a server backend, using the Huggingface Transformers library \cite{wolf-etal-2020-transformers}.
\OD{možná bych tuhle sekci trochu dal víc prominentní, když to má být demo (a ty zbylý zkrátil)}
\RR{Hm... já nevim jestli to nutně má bejt demo... Já v tom spíš vidim Project description paper. Kdyby to byl demo paper tak máme o strnu víc k dispozici. Ale nevim jak se píše demo paper, spíš jsem se chtěl zaměřit na ty technický věci... ale nevim.}
\OD{dalo by se tu mluvit i o ID řádků, historii, možnost stažení}
\RR{Možnost stažení neni ve verzi 1.0. Nevim teď jak vypadá historie ve verzi 1.0. O ID řádků se mluvit dá. Celkově nevím, jestli je cokoliv z toho podstatné.}

% Žere to vstup (typicky scene setting a začátek dialogu), pak to slovo po slově a řádek po řádku generuje pokračování. Vygeneruje 10 replik a vyplivne a čeká na operátora co s tim udělá.

% Sedí u toho Košťák, zadává vstupy, vyhazuje výstupy co se mu nelíběj, případně takové a makové zásahy, až je spokojenej. 

% web app pro toho operátora, na serveru na GPU, 10 replik to generuje asi půl minuty (?)

% \subsection{Low level}

% GPT2 tenaten model

% Neni fajntunovanej, jen tam hackujeme některý věci

% \subsubsection{Script format}

% Many ways how theatre play scripts are often formated. We experimented with various formats, finding that GPT2 works better with some than other. Settled on this:

% \begin{verbatim}
%     A boy and a girl are sitting on a bench.
    
%     Girl: I love you!
    
%     Boy: Thank you.
% \end{verbatim}

% I.e. character names on the same line as their lines, separated by column, empty lines between lines, scenic remarks without any special marks.

\subsection{Resolved Issues}

\subsubsection{Set of characters}

The model does not work with a limited set of characters naturally and tends to forget characters and invent new characters too often.
We resolve this by modifying the next token probability distribution within the GPT-2 model, so that at the start of a new line, only tokens corresponding to character names present in the input prompt are allowed.
We also boost probabilities of characters that have not spoken for a long time.

% Forgets characters, invents characters...

% Detect character names from input prompt, modify the logits so that each line can only start with a character name, and boost probabilities of characters that have not spoken for a long time.

\subsubsection{Repetitiveness}

GPT-2's generation may get stuck in a loop, generating one or several lines again and again.
We managed to resolve this by modifying the hyperparameters of GPT-2, changing repetition penalty from 1.00 to 1.01.
As a backup, we also automatically discard any generated repeated lines and prompt the model to generate another continuing line.

% Forbid repeated lines (identical lines can be repeated max tzwice in a window of 10 consecutive lines.) If a forbidden line is generated, discard it, return the generator to the previous state, and continue generating to get a different variant.

% \subsubsection{Determinism, variability}

% We need to sample outputs to get some variability. We think it is fruitful to have determinism in the sense that identical input leads to identical output. We need to allow the user to throw away a generated line and generate something different.

% We deterministically set the rand seed before generating each line, based on the line ID.
% If generating a new variant for a line, it gets a different ID and so a different rand seed is set and thus a different line is typically generated.
% (Plus can use mechanism of forbidding lines.)

% In case of discarding a forbidden line, we reset the generator state but not the randseed, so that the generator hopefully generates something different this time.

\subsubsection{Limited context}

%Tohle je asi jediný opravdu zajímavý?

% \cite{khandelwal-etal-2018-sharp} conduct perturbation analysis on Neural LM's to investigate how they use context during inference.This paper lays the foundations of what we are trying to apply. As they explore neural caching mechanisms to alleviate the limited context. The idea is to store past computations, with the hope of extrapolating information from the compressed representation of the past. Our surface realization of this idea is textual summarization. 

The variant of the GPT-2 model which we are using has a limit of 1024 subword tokens, within which both the input prompt and the generated output must fit.
The typical solution is to crop the input at the beginning so that it fits
into the window with sufficient space for generating the output. However, this
means forgetting potentially important information from the input prompt and the previously generated
text, which can lead to an unwanted continual topic drift and
also to generating contradictory text;
the text is still locally consistent, but as a whole it may be inconsistent.

To handle this issue, we introduce automated extractive summarization into the process,
hoping that the summarization algorithm will identify the most important
pieces of information to remember.
Whenever the input for GPT-2 (the input prompt + the so far generated script)
exceeds a preset limit of $M=924$ tokens,\footnote{%
%(924 = 1024 - 100)
Most script lines in our setting fit within 100 tokens, so ensuring there is
space for generating at least 100 tokens means that usually the model will
generate a complete line, ending with a newline symbol; in case the generated
line is too long, it is simply cut off once the limit of 1024 tokens is
depleted.}
we summarize the input using TextRank\footnote{%
We use the \texttt{pytextrank} library with minor modifications to reflect the
specific structure of our inputs, so that the algorithm returns $N$ most
important (potentially multi-sentence) \emph{full lines} from the script instead of just $N$ most important
\emph{sentences}.
We set \texttt{limit\_phrases=100}.} \cite{mihalcea_textrank:_2004} before
feeding the input into the GPT-2 model:
\begin{itemize}
    \item We keep all lines within the last $R=250$ tokens from the
        input\footnote{%
        We find the first newline symbol in the last $R$ tokens and keep all
    the lines after it.}
        to ensure local consistency.
    \item We summarize all the preceding lines into $N=5$ lines (while keeping their
        original order) to ensure global consistency.
    \item We concatenate the summary and the kept lines.
    \item If the resulting text is still longer than $M$ tokens, we crop it
        at the beginning to $M$ tokens.
\end{itemize}

\subsubsection{Machine translation}

The GPT-2 model operates on English, while we want to generate a Czech script. We therefore automatically translate the generated script using the CUBBITT~\cite{cubbitt} neural translation model. \OD{The translation result is displayed alongside the English original in THEaiTRobot.}
As the translation tends to discard character names from the lines, we add them by identifying them in the input and translating them independently.

% plus výstup pak ještě cubbittem překládáme do češtiny a hackujeme překlad jmen postav (když se ztratí jméíno poistavym, tak se dopřeloží zvlášť a přilípne se k tomu)

\section{Unresolved issues and future plans}

\subsubsection{Generating a whole play}

The model is not able to generate a long and complex text such as a full theatre play script.
To resolve this, we intend to generate the script hierarchically, first generating a synopsis for the whole play, then expanding it into synopses for individual scenes, and finally generating each scene individually based on its synopsis. This approach is inspired by the work of Fan et al.~\cite{fan_hierarchical_2018,fan_strategies_2019}, who take a similar coarse-to-fine approach to story generation. Our situation is, however, more complex, as we plan to use one more step of the hierarchy.

%\textbf{XXX nějaký citace}
%Fane et al.~\cite{fan2018} Used a LM to first generate a prompt, which was then used as an input to a seq2seq fusion model to generate stories based on the prompt. Their setting was largely limited by the supervised approach, which used a dataset of prompt-story pairs.
%\OD{tohle mi tu asi úplně nesedí – to je vykopírovaný z nějakýho related worku?}
%\DJ{to som napisal motivaciu ktej citacii. Ale teraz je to lepsie preformulovane.}

\subsubsection{Character personalities}

The characters in the play do not seem to have independent personalities in the generated script; the model seems to simply ensure consistency with already generated text, not taking the character names into account. The character personalities thus appear to switch and merge.
We intend to resolve this by learning theatre character embeddings and using them to condition the language model.
We plan to resolve this by clustering our data into several basic character personality types \cite{azab_representing_2019}, then train separate character-aware language models, either by finetuning the GPT-2 model, or by using adapter models \cite{madotto_plug-and-play_2020,wang_k-adapter_2020}.
\OD{ta citace na clustering je trochu off, ale myslim že pro workshop doprí}

%the characters are not self-consistent; the model does not seem to understand that each line belongs to  the  character  whose  name  it  begins  with,  so often the characters are not distinctive but overlap or switch during the dialogue

% characters jsou schizofrení -- zkoušíme se naučit přiřazovadlo postavy k replice

\subsubsection{Dramatic situations}
The text is generated word by word and line by line, whereas human authors of theatre plays typically operate on a more abstract level, such as dramatic situations \cite{polti1921thirty}.\footnote{\url{https://en.wikipedia.org/wiki/The\_Thirty-Six\_Dramatic\_Situations}}
%\textbf{XXXXXX nebo spíš říct že prostě ta hra se odehrává v dramatic situations a netahat do toho jak myslej autoři.}
While there is some work on identifying dramatic turning points \cite{papalampidi_movie_2019,papalampidi_screenplay_2020}, it is too coarse-grained for our application.
We are thus currently annotating a corpus of theatre play scripts with a modified set of dramatic situations, and plan to enhance the tool with this abstraction, either by adding one more layer in the hierarchical setup, or by using special tokens or embeddings to mark dramatic situations in the generated text.

\subsubsection{Machine translation issues}

The MT model we use is tuned for news text, not theatre scripts, and translates each sentence independently. This leads to various issues, including errors in morphological gender (which should pertain to the character), variance in the honorific T–V distinction (which may vary but should be consistent for each pair of characters), and erroneous sentence splitting.
We intend to tackle these issues by using a document-level translation system which takes larger context into account, fine-tuning the model on a corpus of theatre play scripts, and adding various heuristic modifications where necessary.

%MT má mouchy -- hackovat, finetunovat

% \section{Sample From Outputs}

% Ukázat nějakej kousek scény, asi i popsat co je gererated a co je modified a co je jak.
% Ale to se nevejde, leda dát odkaz.

% Možná i zveřejnit někde screencast toho generování a dát sem odkaz.

% Zveřejnit zdrojáky na lindatu a dát sem odkaz.

\section{Conclusion}

We have developed THEaiTRobot 1.0, a tool for interactively generating theatre play scripts.
The tools is based on GPT-2, with several modifications targeting encountered issues.
We have also discussed persisting issues and suggested remedies for a future version.

We used the tool to create the first predominantly machine-generated theatre play script, which is planned for a premiere on 26th February 2021.
Another play, to be generated by an improved version of the tool, is planned for January 2022.

\bibliographystyle{splncs04}
\bibliography{ref}
\end{document}